\documentclass[runningheads]{llncs}

\usepackage{eccv}

\usepackage{eccvabbrv}
\definecolor{LightCyan}{rgb}{0.88,1,1}
\usepackage{multirow}
\usepackage{graphicx}
\usepackage{booktabs}
\usepackage{colortbl}

\usepackage[accsupp]{axessibility}  

\usepackage{hyperref}

\usepackage{orcidlink}

\begin{document}


\title{SPRig: Self-Supervised Pose-Invariant Rigging from Dynamic Mesh Sequences} 

\titlerunning{SPRig}

\author{Ruipeng Wang\inst{1}\thanks{Equal contribution} \and
Langkun Zhong\inst{2}\thanks{Equal contribution} \and
Miaowei Wang\inst{3}}

\authorrunning{R.~Wang et al.}

\institute{University of Pennsylvania, Philadelphia, USA\\
\email{ruipeng@sas.upenn.edu} \and
The University of Hong Kong, Hong Kong, China\\
\email{zhong\_langkun@outlook.com} \and
The University of Edinburgh, Edinburgh, UK\\
\email{s2608314@ed.ac.uk}}

\maketitle

\begin{figure*}[!htb]
    \centering   \includegraphics[width=1\linewidth]{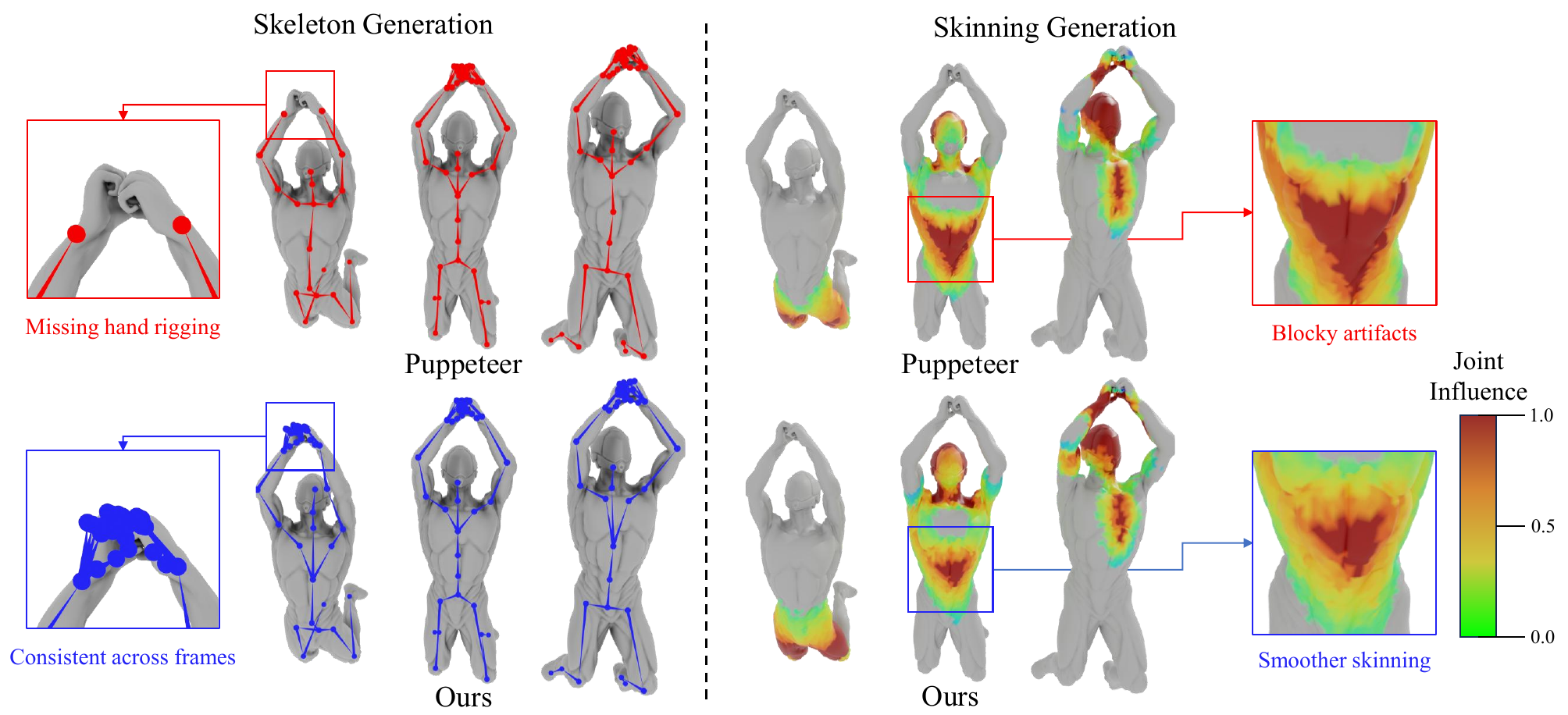}
\vspace{-7mm}
\caption{We propose \textbf{SPRIG}, a fine-tuning framework that enforces cross-frame consistency to learn pose-invariant rigs on existing models. It generates more consistent skeletons and smoother, stable skinning across frames, outperforming static methods such as Puppeteer~\cite{song2025puppeteer}. See the supplementary video for full dynamics.}
\vspace{-10mm}
\label{fig:teaser}
\end{figure*}

\begin{abstract}
State-of-the-art rigging methods typically assume a predefined canonical rest pose. However, this assumption does not hold for dynamic mesh sequences such as DyMesh or DT4D, where no canonical T-pose is available. When applied independently frame-by-frame, existing methods lack pose invariance and often yield temporally inconsistent topologies. To address this limitation, we propose \textbf{SPRig}, a general fine-tuning framework that enforces cross-frame consistency across a sequence to learn pose-invariant rigs on top of existing models, covering both skeleton and skinning generation. For skeleton generation, we introduce novel consistency regularization in both token space and geometry space. For skinning, we improve temporal stability through an articulation-invariant consistency
loss combined with consistency distillation and structural regularization. Extensive experiments show that SPRig achieves superior temporal coherence and significantly reduces artifacts in prior methods, without sacrificing and often even enhancing per-frame static generation quality. The code is available in the supplemental material and will be made publicly available upon publication.
\end{abstract}

\section{Introduction}

3D character rigging, which encompasses skeleton~\cite{chadwick1989layered} and skinning~\cite{lewis2023pose} tuning, remains a fundamental yet time-consuming bottleneck in modern content creation pipelines for games\cite{kavan2007skinning}, film\cite{joshi2007harmonic}, and virtual reality~\cite{marr1978representation}. 

Therefore, many automatic rigging tools~\cite{chu2025humanrig,li2021learning,liu2019neuroskinning,chen2021snarf,liu2025riganything,sun2025armo,yoo2025comprehensive,guo2025make,ma2023tarig,pan2021heterskinnet,song2025magicarticulate,sun2025drive,bian2018automatic,he2025category} have been developed to accelerate this process.  Early automatic rigging methods \cite{baran2007automatic} relied on geometric heuristics and template fitting, which often struggled to generalize across diverse topologies. With the rise of deep learning \cite{lecun2015deep}, \textsc{RigNet} \cite{xu2020rignet} reframed rigging as a learning-based task. More recently, state-of-the-art (SOTA) methods \cite{zhang2025one,song2025puppeteer} have achieved substantial improvements by leveraging powerful Transformer-based \cite{vaswani2017attention} autoregressive models \cite{radford2018improving}, to generate high-quality rigs.

However, these SOTA models are primarily designed for static inputs, specifically requiring a standard T-pose (a standardized neutral posture) for initialization. This assumption is fundamentally incompatible with modern dynamic mesh sequences \cite{li20214dcomplete,wu2025animateanymesh,CAPE:CVPR:20}, including human-designed synthetic animations \cite{objaverse,objaverseXL}, dynamic 3D generative models \cite{wu2025animateanymesh,wang2026bimotion}, 4D video reconstructions \cite{yang2021lasr,goel2023humans,shi2025drive}, and real-world 4D scanning captures \cite{bogo2017dynamic,collet2015high}, which typically lack a fixed canonical reference, either because none is provided or because no corresponding natural pose exists. When applied frame-by-frame independently, these models lack temporal consistency and fail to preserve structural coherence across the sequence. This failure manifests as severe visual artifacts, most notably topological flickering and erratic changes in surface connectivity between frames (\cref{fig:teaser}, Top).

Addressing temporal inconsistency is essential: flickering rigs disrupt animation pipelines and require tedious manual cleanup. Although static rigging datasets are abundant, high-quality labeled dynamic datasets with skeletons and skinning are scarce, necessitating supervision directly from the sequence. Our key insight, rooted in a fundamental graphics principle, is that an animated sequence represents a single object and should have a single, pose-invariant rig. We exploit this as a self-supervised signal: a canonical rig from an anchor frame guides fine-tuning of a pre-trained frame-by-frame model, enforcing consistency across the sequence.

We propose \textbf{SPRig}, a novel fine-tuning framework built upon a frozen-teacher and trainable-student paradigm. ~In this setup, a frozen pre-trained model (the teacher) infers a high-quality rig on a single canonical anchor frame. This prediction then serves as a pose-invariant target to supervise a trainable copy (the student) across all other dynamic frames. By exploiting this self-supervised signal from unlabeled animated mesh sequences, SPRig injects temporal consistency into pretrained static rigging models, making them significantly more robust to pose variations. Specifically, SPRig applies this anchor-based strategy at two stages. For skeleton generation stage, we couple autoregressive token-space constraints with a geometry-space loss that utilizes rigid Procrustes alignment \cite{umeyama2002least} to decouple global motion, forcing the model to replicate the anchor's topology regardless of pose. For skinning generation stage, we establish dense cross-frame correspondences via barycentric tracking, and apply consistency distillation modulated by a soft-support mask to filter out teacher noise. This is further regularized by entropy and geometric proximity priors to guarantee sharp, physically plausible joint assignments. 

Extensive experiments show that our method produces high-quality rigs across dynamic mesh sequences \cite{li20214dcomplete,wu2025animateanymesh} while maintaining and even improving per-frame static rigging performance on challenging out-of-domain datasets. 

In summary, our contributions are as follows:
\begin{itemize}
\item SPRig, a self-supervised fine-tuning framework that enhances temporal stability and pose-invariance in SOTA models using only unlabeled sequences.
\item A set of dual consistency losses for learning pose-invariant skeleton topology and coherent skinning across dynamic mesh frames.
\end{itemize}
\section{Related Work} 
\noindent \textbf{Static automatic rigging.} Early automatic rigging relied on geometric heuristics and template fitting~\cite{amenta1998surface,tagliasacchi2009curve,yan2018voxel,yan2016erosion,au2008skeleton,jin20173d,cao2010point,tagliasacchi2012mean,baran2007automatic,pantuwong2012novel,bhati2013template,orvalho2008transferring,feng2015avatar,pan2009automatic,pantuwong2011fully}. \textsc{Pinocchio}~\cite{baran2007automatic} depends strongly on template quality and struggles with novel topologies. To improve generalization, the field shifted to data-driven approaches. \textsc{RigNet}~\cite{xu2020rignet} framed rigging as an end-to-end learning problem, and subsequent models trained on large artist-created datasets~\cite{deitke2023objaverse,deitke2023objaversexl,lin2025objaverse++,zhang2025texverse} showed strong generalization. With Transformers succeeding in 3D generation~\cite{yu2022point,zhang2022patchformer,lin2021end,zheng2022lightweight,zeng2025renderformer,wu2024point,park2022fast,zhao2021point}, recent SOTAs adopt autoregressive decoding frameworks. Specifically, \textsc{UniRig}~\cite{zhang2025one} formulates rigging as a discrete token sequence generation task, while \textsc{Puppeteer}~\cite{song2025puppeteer} scales up this autoregressive paradigm with a larger model capacity and a significantly expanded, highly diverse training dataset. For skinning, these modern methods typically leverage cross-attention between surface points and bones to robustly assign joint influences. While these methods excel on static assets, they are fundamentally non-temporal. Applying them independently to dynamic sequences leads to severe frame-to-frame inconsistencies (e.g., joint drift, topology flicker) that we target in this work.

\vspace{1mm}
\noindent \textbf{Rigging from dynamic sequences and 4D reconstruction.} A parallel line of research~\cite{anguelov2005scape,james2005skinning,kavan2010fast,mamou2006skinning,de2008automatic,moutafidou2024deep,le2014robust} extracts rigging or motion directly from sequential data. Recent advances extend this to diverse inputs: \textsc{MoRig}~\cite{xu2022morig} extracts motion-aware features to rig 3D point cloud sequences, \textsc{LASR}~\cite{yang2021lasr} recovers articulated shape and kinematics directly from monocular video via differentiable rendering, and \textsc{Reacto}~\cite{song2024reacto} reconstructs articulated neural implicit representations by tracking dense correspondences from casual videos (along with others~\cite{zhang2024magicpose4d,yang2022banmo,wu2023magicpony,yao2025riggs,song2025moda,sun2024ponymation,zhong20254d,sabathier2024animal}). Concurrently, the rapid development of dynamic 3D mesh reconstruction and generation~\cite{wu2025animateanymesh,chen2024ct4d,yuan20244dynamic,bogo2017dynamic} has made high-quality, in-the-wild mesh sequences increasingly obtainable. However, these sequences are typically unlabeled (lacking ground-truth skeletons or skinning). Rather than training a model from scratch to extract rigs from these sequences, \textsc{SPRig} takes a novel perspective: we utilize these abundant, unlabeled dynamic mesh sequences as the perfect self-supervised signal. By enforcing temporal continuity across the sequence, we fine-tune and stabilize the aforementioned static foundation models, marrying powerful static generalization with strict dynamic coherence.

\section{Methodology Overview}
\label{sec:method}
Given an animated mesh sequence $\mathcal{M}=\{\mathbf{M}^k\}_{k=1}^{K}$
with consistent vertex correspondences, we seek a temporally consistent rig: a
skeleton stable in both joint positions and kinematic topology across all
frames (Stage~1, \cref{sec:method:skeleton}), together with skinning weights
that assign each surface point consistently to the same set of joints across
all poses (Stage~2, \cref{sec:method:skinning}).

\smallskip\noindent\textbf{Core principle.}
Both stages share the same teacher--student design \cite{hinton2015distilling}: a \emph{frozen}
pretrained model (teacher) produces a high-quality prediction on anchor
frame~$c$ only; a \emph{trainable} copy (student) is fine-tuned so that its
output on every frame $k$ matches that anchor prediction, yielding pose-invariant
results without retraining from scratch.

\smallskip\noindent\textbf{Anchor frame.}
Both stages are grounded in a single canonical frame $c$, selected
automatically using a geometric criterion. Our key observation is that
the most articulation-free and topologically explicit pose—typically
resembling a T-pose—maximizes the exposed surface area. Rather than
using the conventional first frame \cite{de2008automatic}, which may suffer from self-occlusions or
curled configurations, we search for the most extended pose in the
sequence.  For each frame $\mathbf{M}^k$, we compute its total surface area
$\mathcal{A}(\mathbf{M}^k)$ (sum of triangle areas) and select
$c = \arg\max_{k} \mathcal{A}(\mathbf{M}^k)$ as the anchor frame.
This frame provides a stable canonical reference that reduces the risk
of degenerate joint configurations, and then all meshes are normalized
using its axis-aligned bounding box.

\smallskip\noindent\textbf{Notation.}
We use $J$ joints and $K$ frames.  A skeleton is a pair
$(\mathbf{X}^k\!\in\!\mathbb{R}^{J\times3},\,\mathbf{P}^k\!\in\!\{0,\dots,J\}^J)$
of joint positions and parent indices. We denote $\mathbf{x}^k_j:=[\mathbf{X}^k]_j$ and $p^k_j:=[\mathbf{P}^k]_j$ as the $j$-joint's position and kinematic parent respectively, where $p^k_j\!=\!0$ indicates the root.
The anchor skeleton $(\mathbf{X}^c,\mathbf{P}^c)$ from skeleton generation (Stage~1) is passed as
conditioning input to skinning generation (Stage~2).  We write
$\mathcal{K}:=\{k\mid k\!\neq\!c,\,k\!\in\!\{1,\dots,K\}\}$ for the
non-anchor frame set.

\section{Skeleton Generation}
\label{sec:method:skeleton}

\begin{figure*}[t]
  \centering
  \includegraphics[width=\textwidth,trim=0 0 0 0,clip]{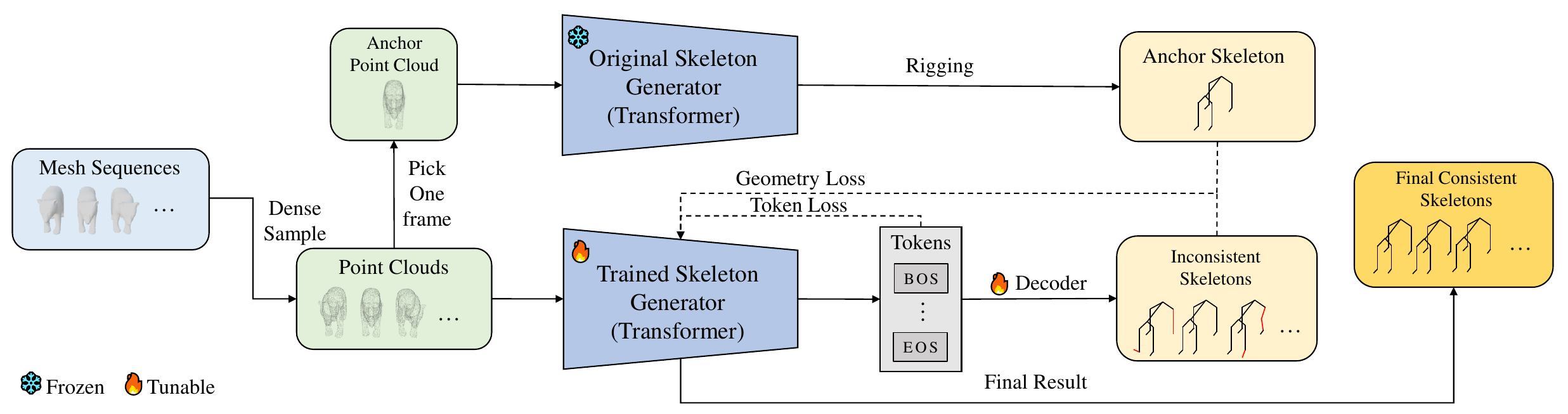}
  \caption{\textbf{Skeleton generation overview.}
Point clouds sampled from the mesh sequence are processed by a shared skeleton generator. The frozen teacher takes only the anchor frame $c$ and produces a canonical token sequence. This sequence is provided as a prefix to the student at every frame. The student is then fine-tuned with token-level and geometry-level consistency losses to generate temporally stable skeletons.}
  \label{fig:skeleton-method}
  \vspace{-6pt}
\end{figure*}

Autoregressive skeleton decoders \cite{zhang2025one,song2025puppeteer,song2025magicarticulate,zhang2026skin} represent a skeleton as a
discrete token stream encoding both geometry and kinematic topology.  While
effective for single-frame prediction, they suffer from temporal
inconsistency: minor shape variations across frames can induce different
tokenizations, manifesting as re-parenting events and positional
jitter\cite{pavllo20193d}.  The root cause is that each frame
is decoded independently, with no cross-frame consistency mechanism. 

We address this with the teacher--student framework (Fig.~\ref{fig:skeleton-method}).
The frozen teacher processes anchor frame~$c$ to generate a stable reference
skeleton and token sequence.  The trainable student is fine-tuned across all
$K$ frames via two consistency losses: token space (\cref{sec:token_consistency})
and geometry space (\cref{sec:geom_consistency}), driving its outputs to
match the anchor reference regardless of pose.

\subsection{Tokenization}
\label{sec:method:skeleton:backbone}

Teacher and student share the same autoregressive decoder $G_\theta$, which
operates on per-frame features $\mathbf{F}^k$ extracted from area-weighted random points \cite{osada2002shape} sampled from mesh $\mathbf{M}^k$. Given
$\mathbf{F}^k$, the decoder generates a token sequence
$\mathbf{t}^k = \{t_i^k\}_{i=1}^{L}$, $L=4J$, factorizing as
$p_\theta(\mathbf{t}^k \mid \mathbf{F}^k)
= \prod_{i=1}^{L} p_\theta(t_i^k \mid t_{<i}^k,\mathbf{F}^k)$.

Tokens $\mathbf{t}^k$ are organized into per-joint quadruples
$\{(t_{j,x}^k,\, t_{j,y}^k,\, t_{j,z}^k,\, t_{j,p}^k)\}_{j=1}^{J}$.
The coordinate tokens $t_{j,x}^k, t_{j,y}^k, t_{j,z}^k \in \{1,\dots,B\}$
encode uniformly quantized Cartesian coordinates in a normalized bounding
box, with $B$ bins.
The parent token $t_{j,p}^k \in \{0,\dots,J\}$ encodes the parent index
$p_j^k$, with $t_{j,p}^k = 0$ indicating the root.
At inference, projecting $\mathbf{t}^k$ yields the skeleton
$(\mathbf{X}^k, \mathbf{P}^k)$.

\subsection{Token Consistency}
\label{sec:token_consistency}

The objective is to make the student approximate the teacher's anchor token
sequence on every frame.  Under \emph{teacher forcing}\cite{bengio2015scheduled,williams1989learning}, the student
at each step $i$ receives the anchor prefix $\hat{t}_{<i}^c$ rather than its
own past outputs, decoupling predictions from its own accumulated errors.  Parent tokens $\{t_{j,p}^k\}$
occupy position $4j$ (1-indexed) of joint $j$'s quadruple.

\smallskip\noindent\textbf{Self-anchor loss} (frame $c$). On frame $c$, we minimize the weighted cross-entropy between the student's
predicted logits ${z}_i^c$ and the teacher's hard token labels
$\hat{t}_i^c$:
\begin{equation}
\label{eq: self_anchor}
  \mathcal{L}_{\mathrm{self}}
  = \frac{\sum_i w_i\,\mathrm{CE}(z_i^c,\, \hat{t}_i^c)}{\sum_i w_i},
\end{equation}
where $w_i$ is empirically set to 5 for parent tokens to prioritize global topological stability and $w_i = 1$ otherwise.

\smallskip\noindent\textbf{Cross-frame loss} ($k \in \mathcal{K}$).
The student receives pose-specific features $\mathbf{F}^k$ but is prefixed
with the anchor sequence $\hat{t}_{<i}^c$, driving its outputs to approximate
the canonical token sequence regardless of pose variations across frames:
\begin{equation}
\label{eq: cross_frame}
  \mathcal{L}_{\mathrm{cross}}
  = \frac{1}{K-1}\sum_{k\in\mathcal{K}}
    \frac{\sum_i w_i\,\mathrm{CE}(z_i^{c,k},\, \hat{t}_i^c)}{\sum_i w_i}.
\end{equation}
The combined token objective
$\mathcal{L}_{\mathrm{token}}
= \lambda_{\mathrm{anc}}\mathcal{L}_{\mathrm{anc}}
+ \lambda_{\mathrm{sym}}\mathcal{L}_{\mathrm{sym}}$,
with $\lambda_{\mathrm{anc}},\lambda_{\mathrm{sym}}\!>\!0$,
trains the student to replicate the anchor token sequence from any pose.

\subsection{Geometry Consistency}
\label{sec:geom_consistency}

Token-level supervision is prone to \emph{exposure bias} \cite{ranzato2015sequence}, where teacher forcing induces a train–test mismatch that accumulates at inference. To mitigate this, we introduce a geometry regularizer $\mathcal{L}_{\mathrm{geom}}$ on the \emph{projected} skeletons, designed to be invariant to global rigid transformations and joint-index permutations.

For each frame $k$, define bone edges
$\mathcal{E}^k := \{(i,j)\mid p^k_j\!=\!i,\;i\!\neq\!j\}$,
bone vectors $\mathbf{v}^k_{ij}\!:=\!\mathbf{x}^k_j\!-\!\mathbf{x}^k_i$, and
midpoints $\mathbf{m}^k_{ij}\!:=\!\tfrac{1}{2}(\mathbf{x}^k_i\!+\!\mathbf{x}^k_j)$.
We factor out global pose via rigid transform     $(\mathbf{R}^k, \mathbf{T}^k)$:
$\mathbf{R}^k$ aligns inertia tensors\cite{gottschalk1996obbtree} through  rigid Procrustes \cite{umeyama2002least}, and
$\mathbf{T}^k = \mu(\mathcal{M}^k) - \mathbf{R}^k\mu(\mathcal{M}^c)$,
where $\mu(\cdot)$ is the centroid and
$\mathcal{M}^k = \{\mathbf{m}^k_{ij}\}_{(i,j)\in\mathcal{E}^k}$.
Let $\mathcal{E}^k_\rho \subseteq \mathcal{E}^k$ be the top-$\rho$ longest
edges. We measure permutation-invariant discrepancies after aligning between each frame and the anchor using three complementary terms, each targeting a distinct failure mode:

\smallskip\noindent\textbf{Bone direction.}
Without directional supervision, token errors can produce limbs pointing the
wrong way.  Let the normalized vector ${\tilde{\mathbf{v}}}^k_{ij} = \mathbf{v}^k_{ij}/\|\mathbf{v}^k_{ij}\|_2$. We define 
$\mathcal{L}_{c\to k} = 1 -
\operatorname{mean}_{(i,j)\in\mathcal{E}^c_\rho}
\max_{(p,q)\in\mathcal{E}^k_\rho}
\langle\mathbf{R}^k{\tilde{\mathbf{v}}}^c_{ij}, {\tilde{\mathbf{v}}}^k_{pq}\rangle,$
and $\mathcal{L}_{k\to c}$ symmetrically. We penalizes inconsistent limb orientations with
$\mathcal{L}_{\mathrm{dir}} =
\tfrac{1}{2}(\mathcal{L}_{c\to k} + \mathcal{L}_{k\to c})$.

\smallskip\noindent\textbf{Bone-length distribution.}
Coordinate-token quantization can cause bone lengths to drift. Let $\ell^k_{(r)}$ be the $r$-th edge length of frame $k$ in ascending order
and $n := \min(|\mathcal{E}^c|, |\mathcal{E}^k|)$. Then we penalize
$\mathcal{L}_{\mathrm{len}} =
\frac{1}{n}\sum_{r=1}^{n}\bigl(\ell^k_{(r)} - \ell^c_{(r)}\bigr)^2$.

\smallskip\noindent\textbf{Joint endpoints.}
The two terms above constrain each bone independently but not their global
spatial arrangement.  Let $\bar{{\mathbf{x}}}^k_i := \mathbf{R}^k{\mathbf{x}}^c_i + \mathbf{T}^k$,
$\mathcal{N}^k := \{[{\mathbf{x}}^k_i |{\mathbf{x}}^k_j]\in\mathbb{R}^6\}_{(i,j)\in\mathcal{E}^k}$, $\bar{\mathcal{N}}^k :=
\{[\bar{{\mathbf{x}}}^k_i | \bar{{\mathbf{x}}}^k_j] \in\mathbb{R}^6\}_{(i,j)\in\mathcal{E}^c}$ where $|$ denotes concatenation, and $d(A,B) := \frac{1}{|A|}\sum_{\mathbf{a}\in A}
\min_{\mathbf{b}\in B}\|\mathbf{a}-\mathbf{b}\|_2^2$ (one-sided Chamfer distance).
 Then we design $\mathcal{L}_{\mathrm{ch}} =
\tfrac{1}{2}[d(\bar{\mathcal{N}}^c, \mathcal{N}^k) +
d(\mathcal{N}^k, \bar{\mathcal{N}}^c)]$
to penalize endpoint mismatch.

The three terms are averaged over non-anchor frames:
\begin{equation}
  \mathcal{L}_{\mathrm{geom}}
  = \frac{1}{K-1}\sum_{k\in\mathcal{K}}
    \bigl(
      \lambda_{\mathrm{dir}}\mathcal{L}_{\mathrm{dir}}
    + \lambda_{\mathrm{len}}\mathcal{L}_{\mathrm{len}}
    + \lambda_{\mathrm{ch}} \mathcal{L}_{\mathrm{ch}}
    \bigr),
  \quad \lambda_{\mathrm{dir}},\lambda_{\mathrm{len}},\lambda_{\mathrm{ch}}\!\geq\!0.
  \label{eq:l_geom}
\end{equation}

\section{Skinning Generation}
\label{sec:method:skinning}

\begin{figure*}[t]
  \centering
  \includegraphics[width=\textwidth,trim=0 0 0 0,clip]{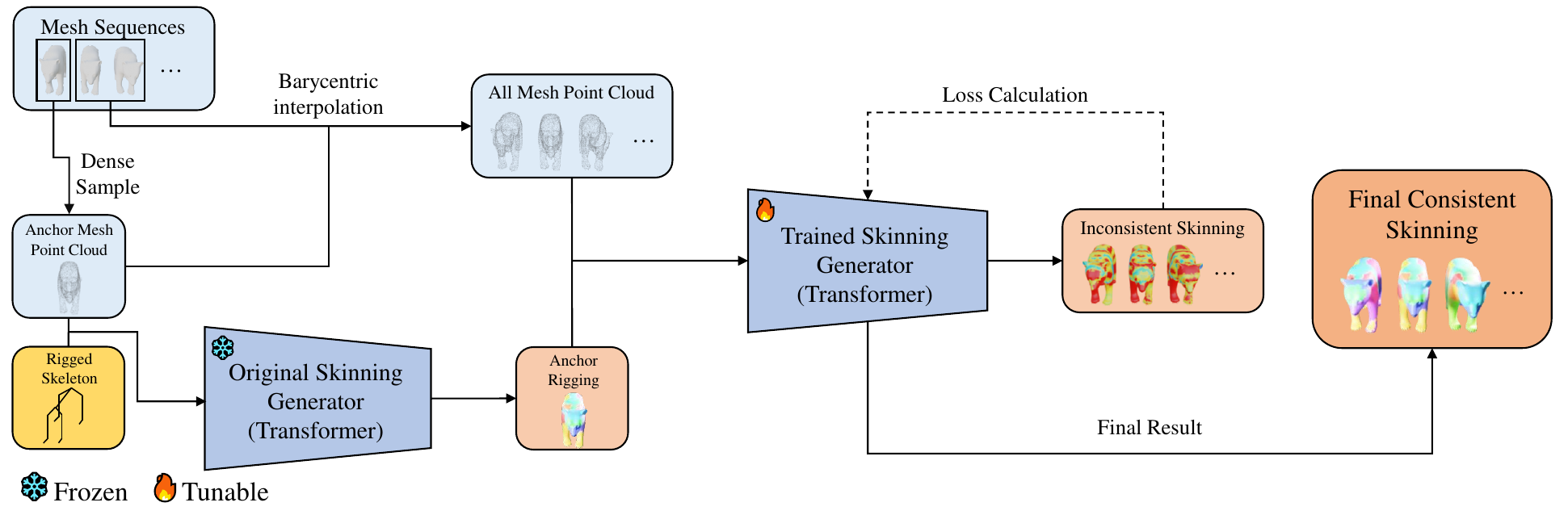}
  \caption{\textbf{Skinning generation overview.}  $N$ surface points are
  sampled from the canonical mesh $\mathbf{M}^c$; the same barycentric
  coordinates track those points across all frames.  The frozen teacher
  predicts skinning weights $\mathbf{W}^c$ on the canonical frame; the
  student $H_\phi$ is fine-tuned so that its predictions $\mathbf{W}^k$ on
  every frame match $\mathbf{W}^c$ via an articulation-invariant consistency
  loss.}
  \label{fig:skinning-method}
  \vspace{-6pt}
\end{figure*}

Given the anchor skeleton $(\mathbf{X}^c, \mathbf{P}^c)$ from Stage~1, we
seek per-point stable skinning-weight matrices $\mathbf{W}^k\!\in\!\mathbb{R}^{N\times J}$
for all frames, where each row encodes how strongly the $N$ surface points
follow each of the $J$ joints.  Na\"{i}ve per-frame prediction yields
inconsistent assignments across poses.  We still apply the teacher--student
strategy: the frozen teacher provides high-quality weights $\mathbf{W}^c$ on
the canonical frame only, which then supervises the student on all frames.

\smallskip\noindent\textbf{Setup.}
We sample $N$ surface points from the canonical $\mathbf{M}^c$ by
triangle-area importance sampling.  For each frame $k$,
applying the same face index and barycentric
coordinates  to the vertices
yields a pose-consistent position $
[\mathbf{E}^k]_i$ for each point $i\in [1,N]$, and the corresponding normal $[\mathbf{N}^k]_i$ is obtained analogously.  We stack these
into $\mathbf{U}^k\!=\![\mathbf{E}^k\,|\,\mathbf{N}^k]\!\in\!\mathbb{R}^{N\times6}$.
From parent index $\mathbf{P}^c$ we derive a valid-joint mask $\mathbf{V}\!\in\!\{0,1\}^J$, where $[\mathbf{V}]_j=1$ iff joint $j\in[1,J]$ is a valid (non-padding) joint, and $[\mathbf{V}]_j=0$ otherwise, and we induce unweighted shortest-path distances on the joint tree $\mathbf{D}\in\mathbb{R}^{J\times J}$; both are shared across frames.

\subsection{Skinning Network}
\label{sec:method:skinning:model}
Trainable students leverage the architecture-agnostic skinning weight generator $H_\phi$ \cite{song2025puppeteer}, which maps per-point queries conditioned on the anchor skeleton $(\mathbf{X}^c,\mathbf{P}^c)$ to per-point weight distributions.
\begin{equation}
  \mathbf{W}^k = H_\phi\bigl(\mathbf{U}^k;\,
    \mathbf{X}^c,\, \mathbf{P}^c,\, \mathbf{V},\, \mathbf{D}\bigr),
  \quad \mathbf{W}^k\in\mathbb{R}^{N\times J},
  \label{eq:skinning_net}
\end{equation}
where each row $[\mathbf{W}^k]_i\in\Delta^{J-1}$, with the simplex
$\Delta^{J-1}\!=\!\{\mathbf{w}\!\in\!\mathbb{R}^J_{\geq0}\mid\sum_j w_j\!=\!1\}$.
$H_\phi$ is a tri-stream Transformer processing point, joint, and shape
features in parallel with topology-aware attention using $\mathbf{D}$ (see details in Supplementary).

\subsection{Canonical-Frame Teacher and Noise Suppression}
\label{sec:method:skinning:teacher}

The frozen teacher is run on the canonical-frame query $\mathbf{U}^c$ to
produce the distillation target:
\begin{equation}
  \hat{\mathbf{W}}^c = H_\phi^{\mathrm{frozen}}\bigl(\mathbf{U}^c;\,
      \mathbf{X}^c,\, \mathbf{P}^c,\, \mathbf{V},\, \mathbf{D}\bigr)
      \in\mathbb{R}^{N\times J}.
  \label{eq:teacher_canonical}
\end{equation}
$\hat{\mathbf{W}}^c$ is fixed and shared across all frames as the pose-invariant
supervision signal.

\smallskip\noindent\textbf{Soft support mask.}
Small entries in $\hat{\mathbf{W}}^c$ reflect negligible joint influence and
introduce noise.  We define a soft support mask
$\mathbf{S}\!\in\![0,1]^{N\times J}$: for each point, the top-$K_s$ valid
joints in $\hat{\mathbf{W}}^c$ receive mask value $1$; remaining valid joints receive a small residual
$\gamma\!\in\!(0,1)$; padding joints are zeroed.  This focuses supervision on
influential joints without discarding gradient signal entirely.

We define two mask-aware operators on any weight matrix
$\mathbf{A}\!\in\!\mathbb{R}^{N\times J}_{\geq0}$: (1) \emph{Masked renormalization}
$\mathcal{R}(\mathbf{A};\mathbf{S})$ restricts $\mathbf{A}$ to the unmasked support:
\begin{equation}
  \mathcal{R}(\mathbf{A};\mathbf{S})
  := \frac{\mathbf{A}\odot\mathbf{1}[\mathbf{S}{>}0]}
          {\sum_j \mathbf{A}_{:,j}\,\mathbf{1}[\mathbf{S}_{:,j}{>}0] + \varepsilon},
  \label{eq:renorm}
\end{equation} where $\varepsilon$ is a small constant added to prevent division by zero. (2) \emph{Masked mean} $[\mathbf{A}]_{\mathbf{S}}$ computes the $\mathbf{S}$-weighted
average over all point--joint pairs:
\begin{equation}
  [\mathbf{A}]_{\mathbf{S}}
  := \frac{N\sum_{i,j} A_{i,j}\,S_{i,j}}
          {\sum_{i,j} S_{i,j}}.
  \label{eq:mavg}
\end{equation}
We write
$\hat{\mathbf{W}}^c_{\mathbf{S}}\!:=\!\mathcal{R}(\hat{\mathbf{W}}^c;\mathbf{S})$
and
$\mathbf{W}^k_{\mathbf{S}}\!:=\!\mathcal{R}(\mathbf{W}^k;\mathbf{S})$
for the masked canonical teacher and student prediction, respectively.

\subsection{Articulation-Invariant Skinning Consistency Loss}
\label{sec:method:skinning:loss}

The student is required to reproduce $\hat{\mathbf{W}}^c_{\mathbf S}$ on every frame.
We enforce this through \emph{consistency distillation}, which directly matches
predictions to $\hat{\mathbf{W}}^c_{\mathbf S}$, and \emph{structural regularization},
which prevents degenerate solutions such as uniformly diffuse weights that minimize
the distillation loss yet lack physical meaning.
The total skinning consistency loss is
$\mathcal{L}_\mathrm{skin}
= \mathcal{L}_\mathrm{cons} + \mathcal{L}_\mathrm{reg}$.

\smallskip\noindent\textbf{Consistency distillation.}
We align the student model's predictions across all frames with $\hat{\mathbf{W}}^c_{\mathbf S}$ using a masked symmetric KL divergence \cite{kullback1951information} and a masked $L_1$ loss. Additionally, to prevent the student from drifting away from the anchor teacher during fine-tuning, we apply an anchor-frame loss $\mathcal{L}_{\mathrm{anchor}}$. The combined consistency loss is $\mathcal{L}_\mathrm{cons} = \lambda_\mathrm{sym}\mathcal{L}_\mathrm{sym} + \lambda_1 \mathcal{L}_1 + \lambda_{\mathrm{anchor}} \mathcal{L}_{\mathrm{anchor}},$
with weights $\lambda_\mathrm{sym}, \lambda_1, \lambda_{\mathrm{anchor}}>0$ controlling the relative weighting of the terms:
\begin{align}
  \mathcal{L}_\mathrm{sym}
    &= \frac{1}{K}\sum_{k=1}^{K}
       \Bigl[
         \mathrm{KL}(\hat{\mathbf{W}}^c_{\mathbf S}\,\|\,\mathbf{W}^k_{\mathbf{S}})
        +\mathrm{KL}(\mathbf{W}^k_{\mathbf{S}}\,\|\,\hat{\mathbf{W}}^c_{\mathbf S})
       \Bigr]_{\!\mathbf{S}}, \label{eq:skl}\\
  \mathcal{L}_{1}
    &= \frac{1}{K}\sum_{k=1}^{K}
       \bigl[|\mathbf{W}^k_{\mathbf{S}} - \hat{\mathbf{W}}^c_{\mathbf S}|\bigr]_{\!\mathbf{S}}, \label{eq:sl1}\\
  \mathcal{L}_{\mathrm{anchor}}
    &= \bigl[|\mathbf{W}^c_{\mathbf{S}} - \hat{\mathbf{W}}^c_{\mathbf S}|\bigr]_{\!\mathbf{S}}. \label{eq:anchor}
\end{align}

\smallskip\noindent\textbf{Structural regularization.} We use two teacher-independent priors to sharpen predictions: (1) The \emph{entropy
penalty} minimizes Shannon entropy \cite{shannon1948mathematical}, encouraging sparse, confident joint
assignments:
\begin{equation}
  \mathcal{L}_\mathrm{ent}
  = -\sum_{k=0}^{K}
    \bigl[\mathbf{W}^k_{\mathbf{S}} \odot \log(\mathbf{W}^k_{\mathbf{S}})
    \bigr]_{\!\mathbf{S}}.
  \label{eq:ent}
\end{equation}
(2) The \emph{geometric proximity prior} prevents sparse and geometrically
implausible assignments: a surface point should primarily follow nearby bones.
For the $i$-th point $[\mathbf{E}^k]_i$ and $j$-joint's bone
$\overline{\mathbf{x}^c_j\mathbf{x}^c_{p^c_j}}$ (from
\cref{sec:geom_consistency}), let $d^k_{i,j}$ denote the shortest Euclidean distance from the point to the segment.  The exponential prior $[
\mathbf{\Pi}^k]_{i,j}\!\propto\!\exp(-\beta d^k_{i,j})$, where $\beta$ is a scaling factor controlling the spatial decay rate,
is averaged over a sliding window $\mathcal{W}_\tau\!\subset\!\{1,\dots,K\}$ of
$\tau$ frames, we empirically set this number to 3:
$\mathbf{\Pi}^k_{\mathcal{W}_\tau}\!=\!\frac{1}{\tau}\sum_{k\in\mathcal{W}_\tau}\mathbf{\Pi}^k$.
The prior loss penalizes deviations from this target:
\begin{equation}
\mathcal{L}_\mathrm{prior} 
= \sum_{k=1}^{K} 
\Bigl[ 
    \mathrm{KL}\Bigl(
        \mathcal{R}(\mathbf{\Pi}^k_{\mathcal{W}_\tau}; \mathbf{S}) \,\|\, 
        \mathbf{W}^k_{\mathbf{S}}
    \Bigr) 
\Bigr]_{\mathbf{S}}.
\label{eq:prior}
\end{equation}
The structural regularization loss is  $\mathcal{L}_\mathrm{reg}
\!=\!\lambda_\mathrm{ent}\mathcal{L}_\mathrm{ent}
\!+\!\lambda_\mathrm{prior}\mathcal{L}_\mathrm{prior}$;
all $\lambda_{(\cdot)}\!\geq\!0$.

\section{Experiment}
\label{sec:experiment}

\providecommand{\TBD}{\textit{TBD}}

\noindent\textbf{Overview.}
We evaluate our method on two tasks: skeleton and skinning generation,
aiming to make SOTA static riggers temporally stable when
applied frame by frame to dynamic mesh sequences. Temporal stability is
measured under an anchor-based protocol, where one frame $c$ is selected
as the canonical anchor and the others are treated as perturbed frames.

\noindent\textbf{Datasets.}
To rigorously evaluate our method, we conduct experiments on two dynamic mesh datasets: (1) we use the standard DeformingThings4D (\textbf{DT4D})~\cite{li20214dcomplete} dataset, constructing a 125-sequence validation set by randomly sampling one sequence from each object category.
(2) To verify that our framework scales to diverse in-the-wild topologies, we introduce a curated subset of \textbf{DyMesh}~\cite{wu2025animateanymesh}, comprising 1007 high-quality motion sequences with extreme deformations and varied skeletal structures.

\noindent\textbf{Implementation Details.}
We instantiate SPRig by fine-tuning the pretrained Puppeteer models, though our method is agnostic to the specific Transformer architecture (see Supp. for details). Specifically, we freeze the geometry point cloud encoders and only update the Transformer decoders using self-supervised signals from the dynamically discovered canonical anchors. All experiments are implemented in PyTorch \cite{paszke2019pytorch} and conducted on a single NVIDIA A100 GPU. We utilize the AdamW \cite{loshchilovdecoupled} optimizer for both tasks: the skeleton network is trained with a learning rate of $3 \times 10^{-6}$ and a batch size of $20$, while the skinning network uses a learning rate of $1 \times 10^{-5}$ and a batch size of $12$ with automatic mixed  \cite{micikevicius2018mixed}. During fine-tuning, both models converge efficiently within 10 to 12 epochs. The entire training process takes about 20 hours to finish.

\subsection{Skeleton Generation}

\noindent\textbf{Metrics.}
We report: 
(1)  the pairwise joint distance deviation (\textbf{PJDD}), which measures geometric jitter over time by quantifying the drift in all pairwise joint distances relative to the canonical anchor after rigid alignment (see Suppl. for details);
(2) \textbf{GSD}~\cite{chung1997spectral}, which evaluates global structural consistency based on spectral graph theory. Both PJDD and GSD serve as our dynamic metrics evaluated on the DT4D and DyMesh test splits; 
(3) \textbf{CD-J2J}, \textbf{CD-J2B}, and \textbf{CD-B2B}~\cite{xu2020rignet}, which serve as representative static quality metrics, which are computed on \textbf{Articulation-XL 2.0}~\cite{song2025magicarticulate}, \textbf{ModelsResource}~\cite{xu2020rignet}, and \textbf{Diverse-pose}~\cite{song2025puppeteer} test sets for out-of-domain generalization.

\begin{figure}[t!]
    \centering
    \begin{minipage}[b]{0.48\linewidth}
        \centering
        \includegraphics[width=\linewidth]{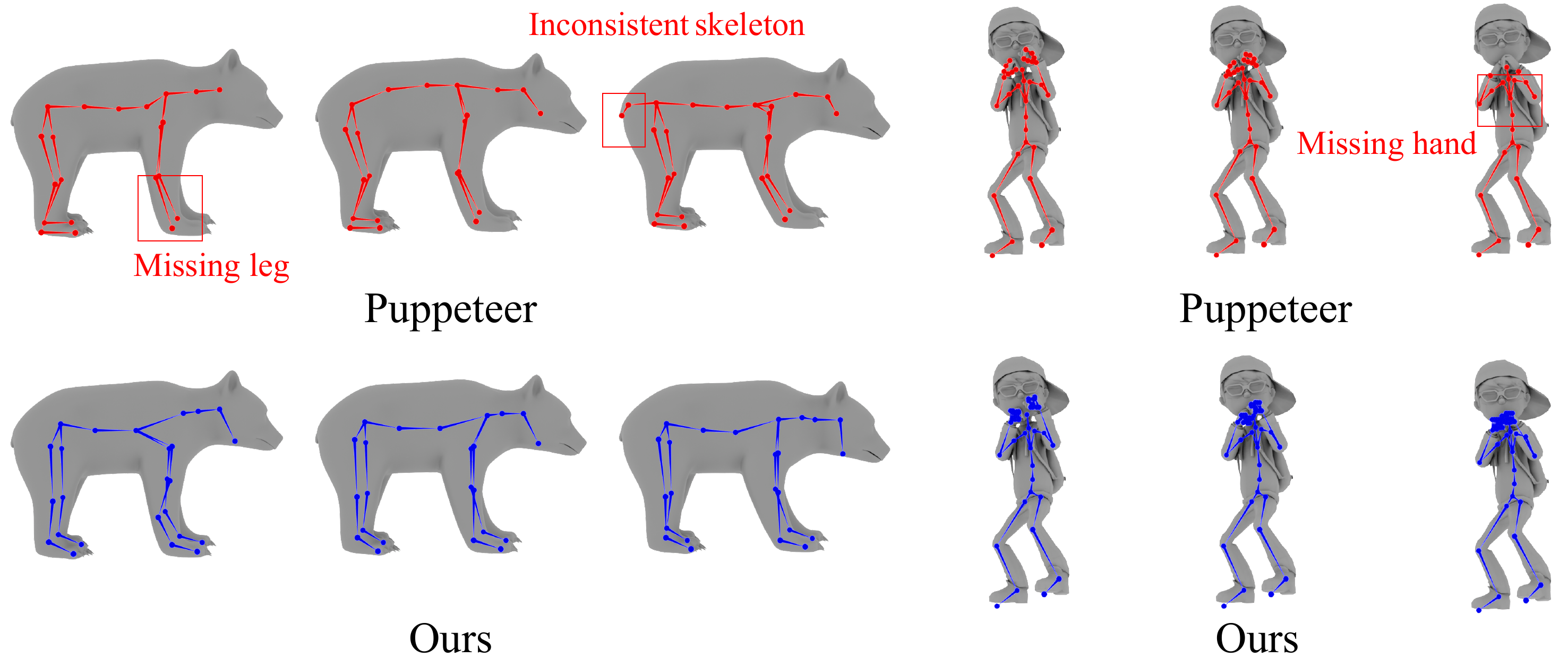}
    \end{minipage}
    \hfill
    \begin{minipage}[b]{0.48\linewidth}
        \centering
        \includegraphics[width=\linewidth, trim=0 0 0 0, clip]{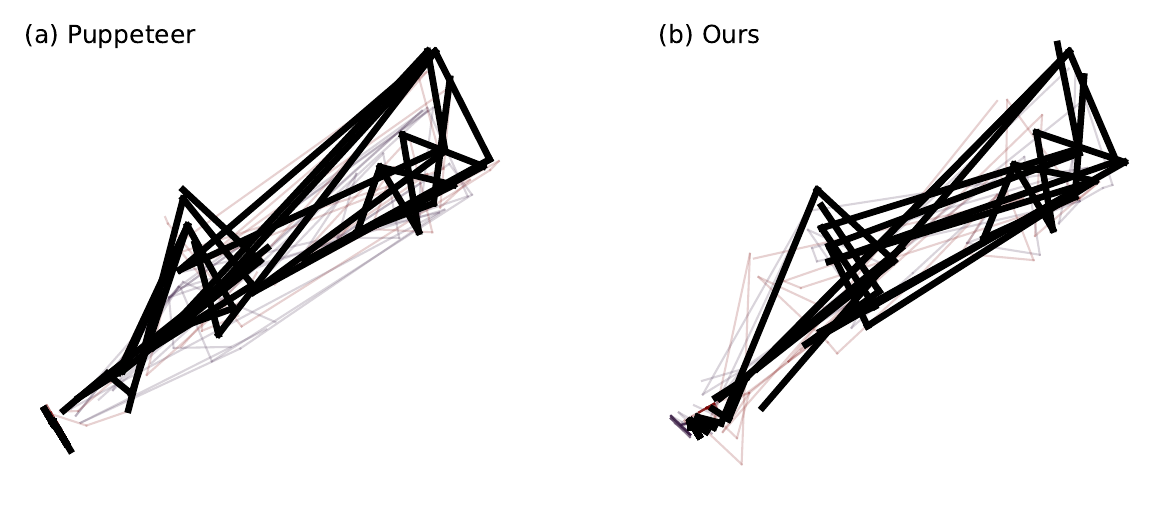}
    \end{minipage}
    
    \vspace{2mm} 
    
    \begin{minipage}[t]{0.48\linewidth}
    \caption{
            \textbf{Skeleton qualitative comparisons.}
            Top (red): the Puppeteer often misses or distorts structures (e.g., missing leg on the bear and missing hand on the human), and exhibits inconsistent skeleton topology between frames.
            Bottom (blue): our method produces more temporally stable and complete skeletons across frames.
             By enforcing cross-frame consistency, SPRig effectively prevents this topological drift and preserves structural integrity within sequences.
        }
        \label{fig:qual_skeleton_bear_human}
    \end{minipage}
    \hfill
    \begin{minipage}[t]{0.48\linewidth}
    \caption{
        \textbf{Skeleton temporal jitter.} To isolate structural jitter from global motion, we rigidly align all predicted skeletons in a sequence to a canonical anchor frame using Procrustes analysis. Puppeteer shows noticeable geometric jitter and inconsistent bone lengths, whereas our method learns a pose-invariant prior that keeps skeletons tightly clustered around the anchor, improving temporal stability. Thick black lines denote the canonical anchor, and thin translucent colored lines denote other frames.
    }
        \label{fig:skeleton_trajectory}
    \end{minipage}
    \vspace{-4mm}
\end{figure}


\begin{table}[!htb]
\centering
\small
\caption{\textbf{Dynamic quantitative comparisons.} Left: Skeleton topology generation. Right: Skinning weight generation. Our method substantially improves both skeleton and skinning metrics across DT4D and DyMesh datasets.}
\label{tab:dynamic_combined}
\resizebox{\linewidth}{!}{%
\begin{tabular}{l | cc | cc | cc | cc}
\toprule
\multirow{3}{*}{Model} 
& \multicolumn{4}{c|}{Skeleton Generation} 
& \multicolumn{4}{c}{Skinning Generation} \\

\cmidrule(r){2-5} \cmidrule(r){6-9}
& \multicolumn{2}{c|}{DT4D}
& \multicolumn{2}{c|}{DyMesh}
& \multicolumn{2}{c|}{DT4D}
& \multicolumn{2}{c}{DyMesh} \\
& PJDD $\downarrow$ & GSD $\downarrow$ & PJDD $\downarrow$ & GSD $\downarrow$ 
& $\mathrm{L_1}$ Erro $\downarrow$ & LBS RMSE $\downarrow$ & $\mathrm{L_1}$ Error $\downarrow$ & LBS RMSE $\downarrow$ \\
\midrule
UniRig & 15.76 & 0.060 & 19.56 & 0.071 & 1310.31 & 0.007576 & 1430.43 & 0.007763 \\
Puppeteer & 17.46 & 0.062 & 16.53 & 0.068 & 1328.80 & 0.007560 & 1379.32 & 0.007835 \\
\cellcolor{LightCyan}\textbf{Ours} & \cellcolor{LightCyan}\textbf{0.68} & \cellcolor{LightCyan}\textbf{0.056} & \cellcolor{LightCyan}\textbf{0.72} & \cellcolor{LightCyan}\textbf{0.054} 
& \cellcolor{LightCyan}\textbf{982.35} & \cellcolor{LightCyan}\textbf{0.007552} & \cellcolor{LightCyan}\textbf{936.58} & \cellcolor{LightCyan}\textbf{0.007447} \\
\bottomrule
\end{tabular}%
}
\end{table}

\begin{table}[!htb]
\centering
\caption{\textbf{Skeleton static prediction comparisons.} Our method maintains strong per-frame fidelity on diverse out-of-domain static datasets.}
\label{tab:skeleton_static_quality}
\resizebox{\textwidth}{!}{%
\begin{tabular}{l | ccc | ccc | ccc}
\toprule
\multirow{2}{*}{Model} & \multicolumn{3}{c|}{Articulation-XL 2.0} & \multicolumn{3}{c|}{ModelsResource} & \multicolumn{3}{c}{Diverse-Pose} \\
& CD-J2J $\downarrow$ & CD-J2B $\downarrow$ & CD-B2B $\downarrow$ & CD-J2J $\downarrow$ & CD-J2B $\downarrow$ & CD-B2B $\downarrow$ & CD-J2J $\downarrow$ & CD-J2B $\downarrow$ & CD-B2B $\downarrow$ \\
\midrule
UniRig & 0.0331 & 0.0261 & 0.0218 & 0.0396 & 0.0302 & 0.0257 & 0.0325 & 0.0257 & 0.0208 \\
Puppeteer & 0.0311 & 0.0237 & 0.0198 & 0.0377 & 0.0280 & 0.0241 & 0.0251 & 0.0199 & 0.0160 \\
\cellcolor{LightCyan}\textbf{Ours} & \cellcolor{LightCyan}\textbf{0.0270} & \cellcolor{LightCyan}\textbf{0.0213} & \cellcolor{LightCyan}\textbf{0.0188} & \cellcolor{LightCyan}\textbf{0.0325} & \cellcolor{LightCyan}\textbf{0.0258} & \cellcolor{LightCyan}\textbf{0.0235} & \cellcolor{LightCyan}\textbf{0.0191} &\cellcolor{LightCyan}\textbf{0.0178} & \cellcolor{LightCyan}\textbf{0.0125} \\
\bottomrule
\end{tabular}%
}
\end{table}


\begin{figure}[!htb]
    \centering
    \begin{minipage}[c]{0.35\linewidth} 
        \makeatletter\def\@captype{table}\makeatother
        \caption{
        \textbf{Quantitative skeleton ablations.}  Removing the token loss ($\mathcal{L}_{\mathrm{token}}$), geometric loss ($\mathcal{L}_{\mathrm{geom}}$), or the hierarchical parent weight ($w_i=5$) significantly degrades both geometric stability (PJDD) and structural consistency (GSD). All variants are evaluated on the DT4D dataset.}
        \label{tab:skeleton_ablation_vertical}
        \centering
        \resizebox{1\linewidth}{!}{%
        \begin{tabular}{lcc}
        \toprule
        Model & PJDD $\downarrow$ & GSD $\downarrow$ \\
        \midrule
        \cellcolor{LightCyan}\textbf{Default} & \cellcolor{LightCyan}\textbf{0.68}  & \cellcolor{LightCyan}\textbf{0.056} \\
        w/o. $\mathcal{L}_{\mathrm{token}}$    & 1.20  & 0.047 \\
        w/o. $\mathcal{L}_{\mathrm{geom}}$     & 1.38  & 0.067 \\
        w/o. $w_i=5$                            & 1.36  & 0.073 \\
        \bottomrule
        \end{tabular}%
        }
    \end{minipage}
    \hfill
    \begin{minipage}[c]{0.58\linewidth} 
        \makeatletter\def\@captype{figure}\makeatother
        \includegraphics[width=\linewidth]{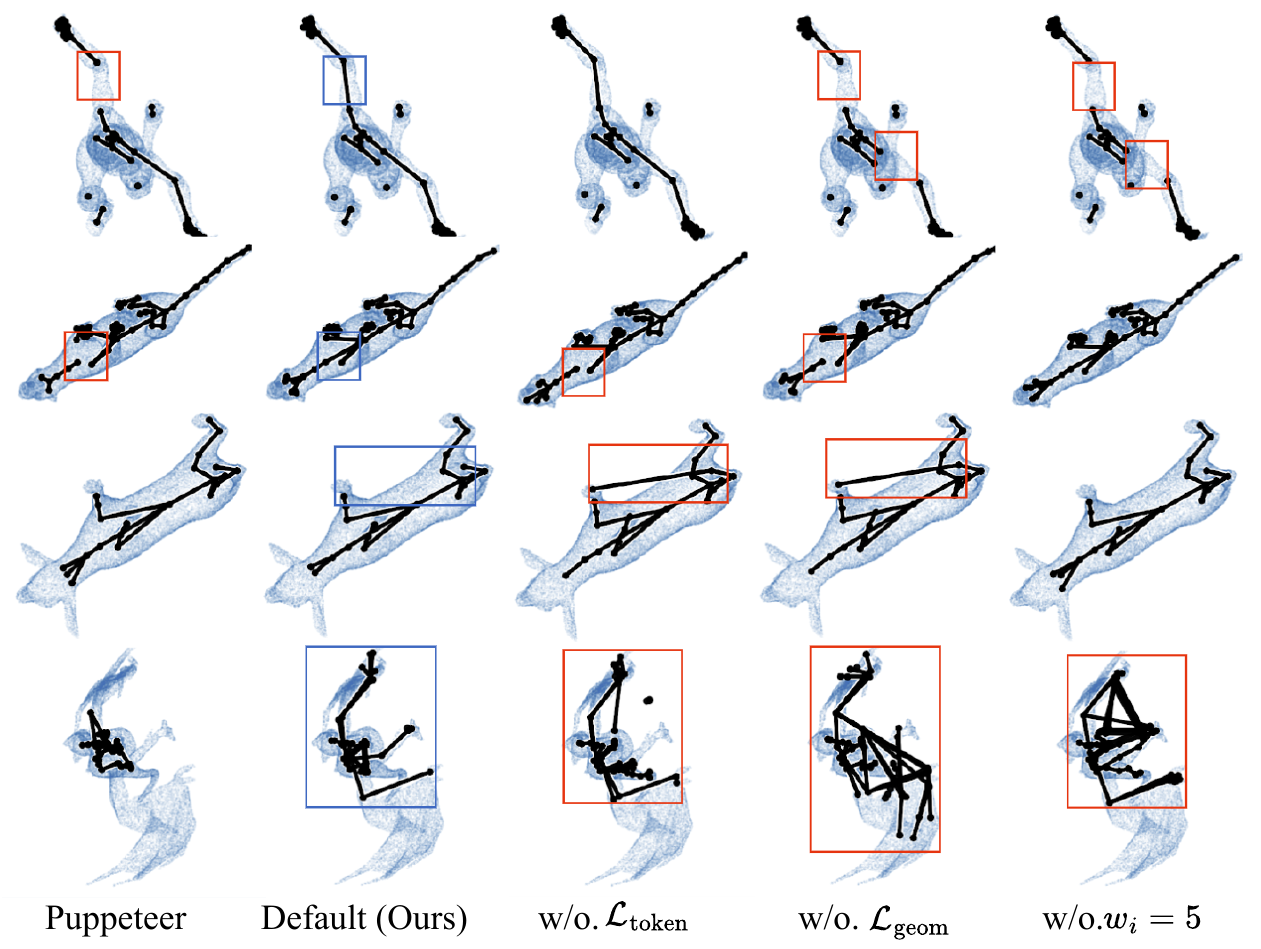}
        \caption{
            \textbf{Qualitative skeleton ablations.} 
      Removing $\mathcal{L}_{\mathrm{token}}$ often omits critical limbs; removing $w_i=5$ breaks hierarchical connectivity, causing chaotic structures; and omitting $\mathcal{L}_{\mathrm{geom}}$ results in poor geometric alignment. Our default model consistently recovers complete skeletons.”
        }
        \label{fig:skeleton_ablation_qual}
    \end{minipage}
    \vspace{-4mm}
\end{figure}

\noindent\textbf{Results.}
SPRig (see \cref{tab:dynamic_combined}, Left) drastically suppresses geometric jitter compared to baselines, reducing PJDD  from 17.46 to 0.68 on DT4D while maintaining superior structural consistency (GSD).  ~Importantly, this substantial improvement in temporal stability does not compromise per-frame fidelity; as reported in \cref{tab:skeleton_static_quality}, SPRig preserves and even improves baseline static quality on all evaluated datasets, demonstrating strong out-of-domain generalization. Besides, our visual results (\cref{fig:qual_skeleton_bear_human} and \cref{fig:skeleton_trajectory}) show that SPRig effectively eliminates missing joints and severe inter-frame flicker, producing structurally complete and smoothly moving rigs across highly diverse sequences. This reinforces our conclusion that grounding the generation process to a canonical anchor enables the model to internalize a robust, pose-invariant structural prior.


\begin{figure}[t!]
    \centering
    \begin{minipage}[t]{0.48\linewidth}
        \vspace{0pt} 
        \makeatletter\def\@captype{table}\makeatother
        \caption{
            \textbf{Quantitative ablation of anchor frame.} Comparing the  first frame ($c=1$) with our automatic discovery ($c=\arg\max_{k}\mathcal{A}$), our approach improves both skeleton and skinning metrics on the DT4D dataset.
        }
        \label{tab:anchor_definition}
        \centering
        \resizebox{0.8\linewidth}{!}{%
        \begin{tabular}{lcc}
        \toprule
        Metric & $c=1$ & \cellcolor{LightCyan}\textbf{Default} \\
        \midrule
        \multicolumn{3}{c}{Skeleton Generation} \\
        \midrule
        PJDD $\downarrow$       & 0.68   & \cellcolor{LightCyan}\textbf{0.59} \\
        GSD $\downarrow$        & 0.056  & \cellcolor{LightCyan}\textbf{0.046} \\
        CD-B2B $\downarrow$     & 0.0188 & \cellcolor{LightCyan}\textbf{0.0175} \\
        \midrule
        \multicolumn{3}{c}{Skinning Generation} \\
        \midrule
        $\mathrm{L_1}$ Error $\downarrow$   & 1001.56  & \cellcolor{LightCyan}\textbf{982.35} \\
        LBS RMSE $\downarrow$   & 0.007558 & \cellcolor{LightCyan}\textbf{0.007552} \\
        \bottomrule
        \end{tabular}%
        }
    \end{minipage}
    \hfill
    \begin{minipage}[t]{0.48\linewidth}
        \vspace{0pt} 
        \makeatletter\def\@captype{figure}\makeatother
        \includegraphics[width=\linewidth]{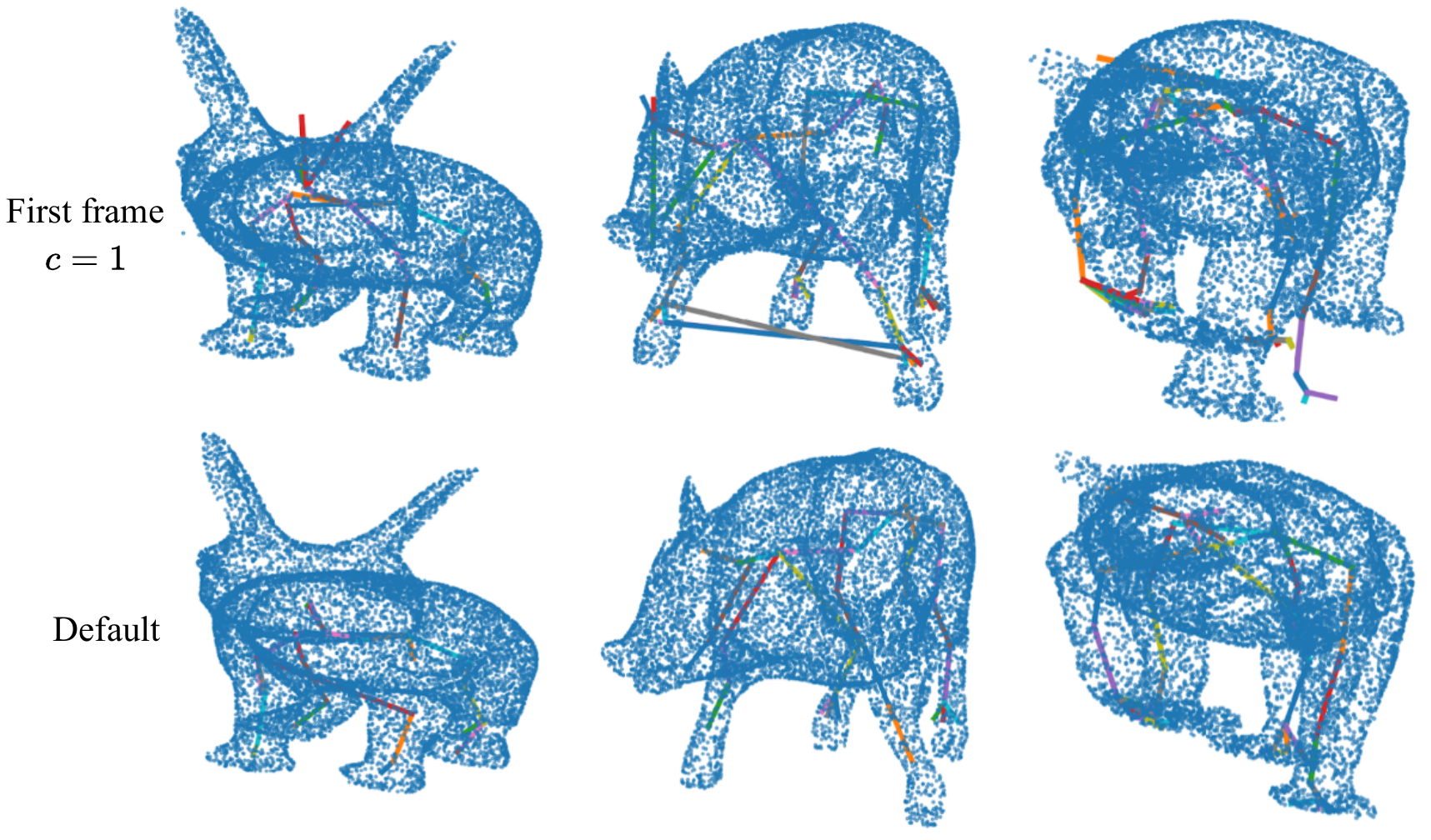}
        \caption{
            \textbf{Qualitative ablation of anchor frame.} 
            Conventional first frame ($c{=}1$) fails on severe initial self-occlusions (e.g., squatted poses), yielding corrupted skeletons. Our dynamic selection ($c = \arg\max_{k} \mathcal{A}$) automatically identifies the optimal rest pose for clean rigging.
        }
        \label{fig:anchor_discovery}
    \end{minipage}
    \vspace{-4mm}
\end{figure}
\noindent\textbf{Ablations.}
We ablate key components of our dual-consistency training objective (see \cref{tab:skeleton_ablation_vertical}) and the parent-token weighting strategy ($w_i=5$ in \cref{sec:token_consistency}). Removing either the geometry-space loss or the token-space loss substantially worsens temporal stability, confirming that both constraints are critical. Similarly, reverting to standard cross-entropy (with all $w_i=1$) by removing the parent-token weighting also degrades stability, highlighting that emphasizing root tokens is essential for stabilizing skeleton topology and reducing geometric jitter. Additionally, we assess the effect of our dynamic anchor selection (see \cref{tab:anchor_definition}, Top, and \cref{fig:anchor_discovery}): automatically identifying the optimal rest pose, instead of defaulting to the first frame, is crucial for preventing initial topological corruption and further reducing geometric jitter.

\subsection{Skinning Generation}

\begin{figure*}[t!]
    \centering
        \vspace{2mm}
    \includegraphics[width=\textwidth, trim=0 0 0 0, clip]{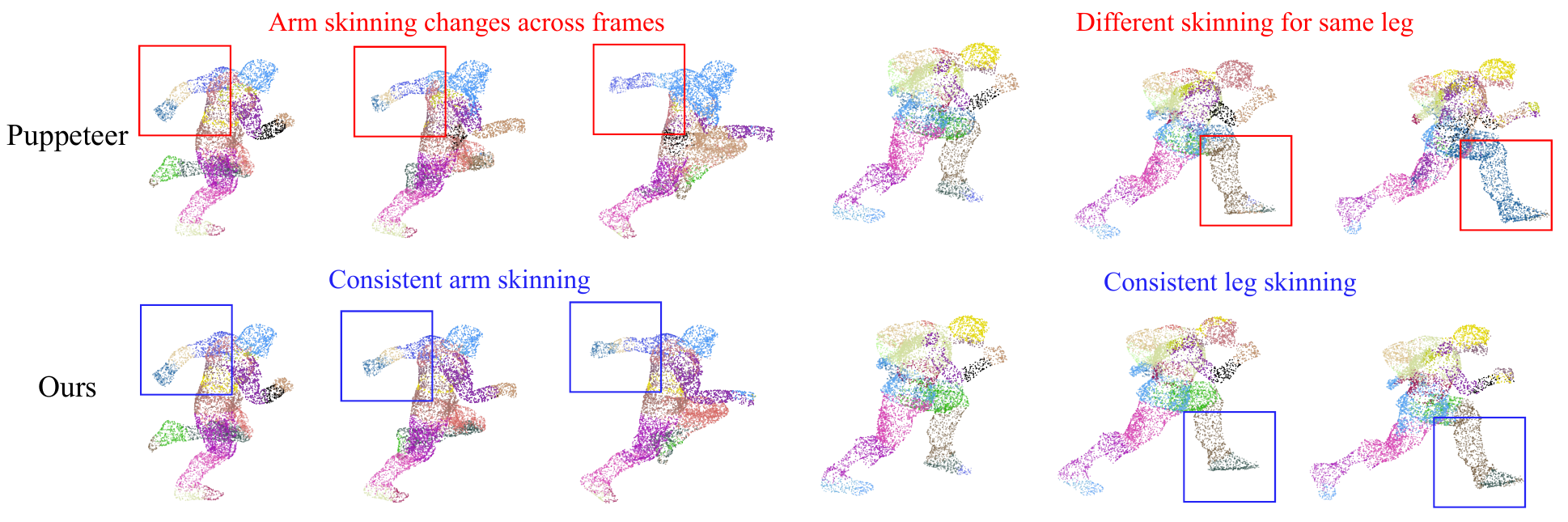}
    \caption{
\textbf{Skinning qualitative comparisons.} Results on running human sequences from DT4D are visualized, with each vertex colored by the joint with the maximum predicted skinning influence. Our method produces temporally consistent assignments for arms and legs across frames (blue boxes), whereas Puppeteer shows severe flickering, with dominant joint assignments switching between frames (red boxes).
    } 
    \label{fig:qual_skinning_running_human}
    \vspace{-3mm}
\end{figure*}

\noindent\textbf{Metrics.}
Similar to the skeleton evaluation, we report both dynamic and static metrics: (1) Temporal \textbf{$\mathbf{L_1}$ Error}: Unlike VIBE’s \cite{kocabas2020vibe} acceleration error, which measures second-order joint acceleration, we compute the first-order variation of skinning weights to directly capture temporal flicker and inconsistencies in predicted attachments across frames (see Suppl. for details). (2) \textbf{LBS RMSE}: Measures the accuracy of static surface deformation. We drive the canonical mesh using the predicted rigs via standard Linear Blend Skinning \cite{magnenat1989joint} and compute the vertex-wise reconstruction error against the ground truth. (3) \textbf{Precision}, \textbf{Recall}, and \textbf{Avg. $\mathbf{L_1}$ Dist}. \cite{xu2020rignet}: Standard metrics for static prediction quality. To verify out-of-domain single-frame robustness, these metrics are also evaluated on Articulation-XL 2.0 \cite{song2025magicarticulate}, ModelsResource \cite{xu2020rignet}, and the challenging Diverse-pose \cite{song2025puppeteer} datasets.

\begin{figure}[t!]
  \centering
  \includegraphics[width=\linewidth ,trim=0 0 0 0,clip]{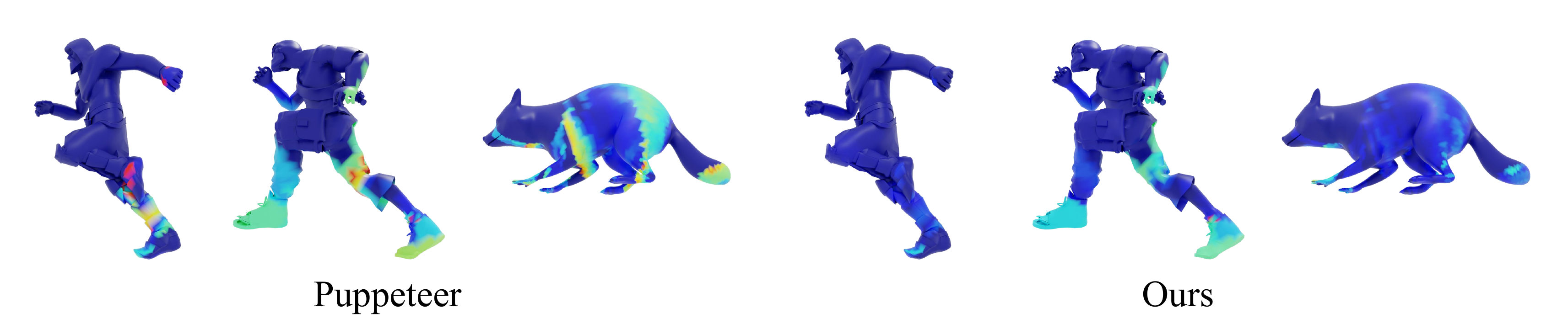}
  \caption{
    \textbf{Temporally consistent skinning.}
We visualize per-vertex inconsistencies between perturbed-frame skinning predictions and the canonical anchor teacher using the $\mathrm{L}_1$ deviation of predicted weights (blue: low error; red: high error). Puppeteer (Left) exhibits large high-error regions, indicating strong temporal flicker, whereas our method (Right) largely suppresses these errors, consistent with the results in \cref{tab:dynamic_combined}, Right. Additional discussion is provided in the Supplement.}
\label{fig:error_maps}
\end{figure}

\noindent\textbf{Results.}
Our method (see \cref{tab:dynamic_combined}, Right) significantly reduces the temporal $\mathrm{L_1}$ error compared to the static baselines, demonstrating that our fine-tuning effectively mitigates inter-frame flicker in skinning assignments. At the same time, the LBS RMSE is consistently improved over both baselines, indicating that the substantial gains in temporal stability do not compromise—and in fact enhance—the deformation fidelity for skinning generation. Moreover, our fine-tuning strategy maintains competitive performance on out-of-domain static datasets (see \cref{tab:skinning_static_quality}) and even surpasses all baselines on the Diverse-Pose challenge dataset. Visually, our method (see \cref{fig:qual_skinning_running_human,fig:error_maps}) produces highly consistent vertex-to-joint skinning attachments across frames and effectively suppresses per-vertex temporal deviations.

\begin{table}[!htbp]
\centering
\caption{\textbf{Skinning static prediction comparisons.} Our method achieves competitive per-frame fidelity on Articulation-XL 2.0 and ModelsResource, while consistently outperforming the baseline on the challenging Diverse-pose dataset, demonstrating greater robustness to extreme articulations.}
\label{tab:skinning_static_quality}
\resizebox{\textwidth}{!}{%
\begin{tabular}{l | ccc | ccc | ccc}
\toprule
\multirow{2}{*}{Model} & \multicolumn{3}{c|}{Articulation-XL 2.0} & \multicolumn{3}{c|}{ModelsResource} & \multicolumn{3}{c}{Diverse-Pose} \\
& Precision $\uparrow$ & Recall $\uparrow$ & Avg. $\mathrm{L}_1$ Dist. $\downarrow$ & Precision $\uparrow$ & Recall $\uparrow$ & Avg. $\mathrm{L}_1$ Dist. $\downarrow$ & Precision $\uparrow$ & Recall $\uparrow$ & Avg. $\mathrm{L}_1$ Dist. $\downarrow$ \\
\midrule
RigNet & 0.737 & 0.661 & 0.729 & 0.657 & 0.802 & 0.707 & 0.747 & 0.654 & 0.746 \\
Puppeteer & \textbf{0.876} & 0.740 & \textbf{0.335} & \textbf{0.797} & 0.816 & \textbf{0.443} & 0.836 & 0.722 & 0.405 \\
\cellcolor{LightCyan}\textbf{Ours} & \cellcolor{LightCyan}0.863 & \cellcolor{LightCyan}\textbf{0.745} & \cellcolor{LightCyan}0.355 & \cellcolor{LightCyan}0.732 & \cellcolor{LightCyan}\textbf{0.883} & \cellcolor{LightCyan}0.462 & \cellcolor{LightCyan}\textbf{0.842} & \cellcolor{LightCyan}\textbf{0.734} & \cellcolor{LightCyan}\textbf{0.378} \\
\bottomrule
\end{tabular}%
}
\end{table}

\begin{figure}[t!]
    \centering
    \begin{minipage}[t]{0.45\linewidth}
        \vspace{0pt} 
        \makeatletter\def\@captype{table}\makeatother
        \caption{\textbf{Quantitative skinning ablations.} 
        Removing any loss term significantly degrades temporal stability ($\mathrm{L}_1$ Error) and fidelity (LBS RMSE). Models trained for 24 epochs. All variants are evaluated on the
DT4D dataset.}
        \label{tab:skinning_ablation_vertical}
        \centering
        \resizebox{0.9\linewidth}{!}{%
        \begin{tabular}{lcc}
        \toprule
        Model & $\mathrm{L}_1$ Error $\downarrow$ & LBS RMSE $\downarrow$ \\
        \midrule
        \cellcolor{LightCyan}\textbf{Default} & \cellcolor{LightCyan}\textbf{982.35} & \cellcolor{LightCyan}\textbf{0.007552} \\
        w/o. $\mathcal{L}_{\mathrm{sym}}$    & 3897.24 & 0.016587 \\
        w/o. $\mathcal{L}_{1}$               & 4007.13 & 0.023254 \\
        w/o. $\mathcal{L}_{\mathrm{anchor}}$ & 4006.53 & 0.023305 \\
        w/o. $\mathcal{L}_{\mathrm{ent}}$    & 4007.74 & 0.023311 \\
        w/o. $\mathcal{L}_{\mathrm{prior}}$  & 4006.78 & 0.023347 \\
        \bottomrule
        \end{tabular}%
        }
    \end{minipage}
    \hfill
    \begin{minipage}[t]{0.48\linewidth}
        \vspace{0pt} 
        \makeatletter\def\@captype{figure}\makeatother
        \includegraphics[width=\linewidth]{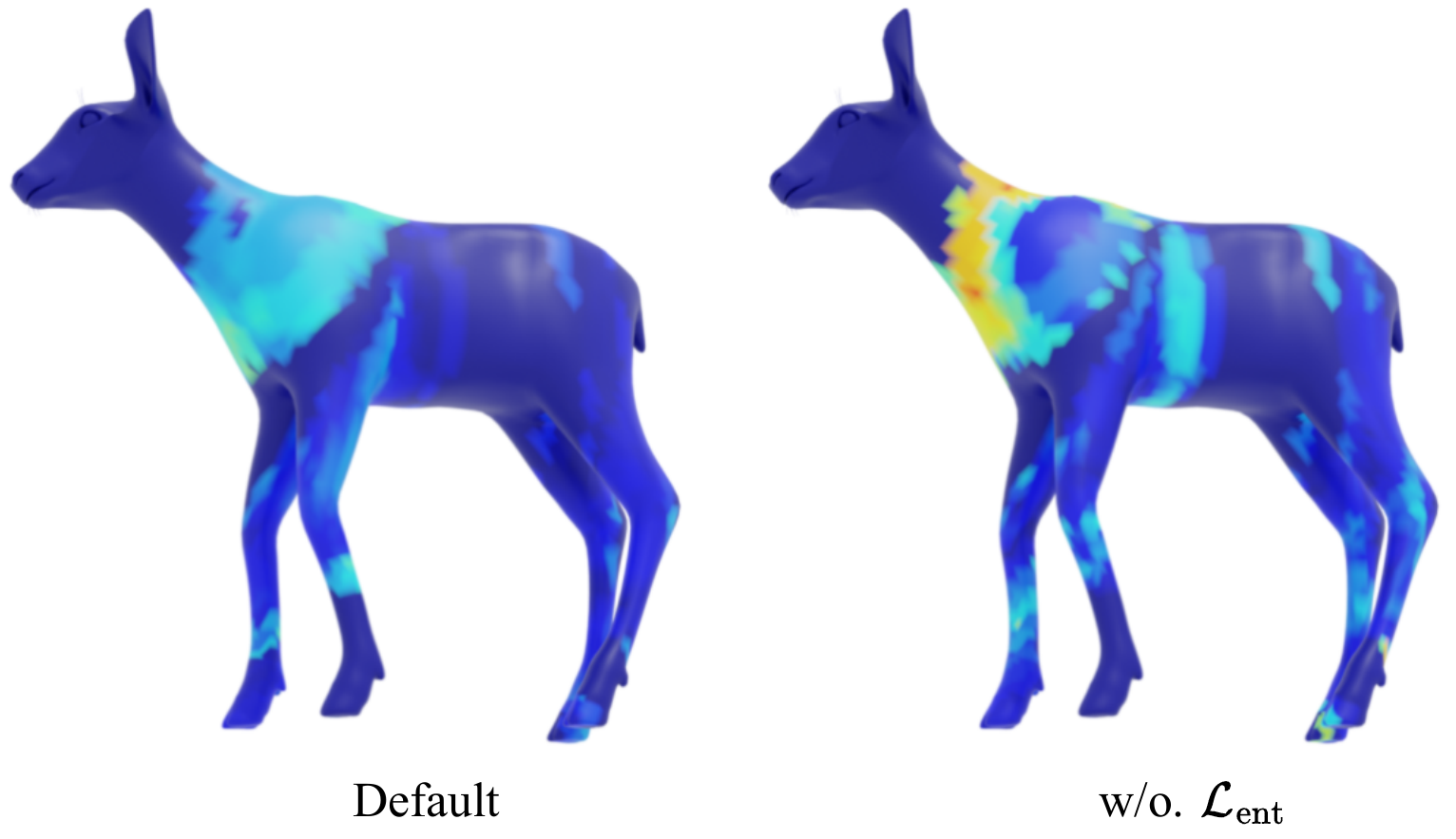}
        \caption{
            \textbf{Qualitative $\mathcal{L}_{\mathrm{ent}}$ ablation.} Per-vertex Shannon entropy is visualized (blue: sharp; red: diffuse). Our full model (Left) produces sparse, confident assignments, whereas removing $\mathcal{L}_{\mathrm{ent}}$ yields ambiguous weights, reducing stability. 
        }
        \label{fig:skinning-ablation-entropy}
    \end{minipage}
    \vspace{-4mm}
\end{figure}

\begin{figure*}[t!]
    \centering
    \includegraphics[width=\linewidth, trim=0 0 0 0, clip]{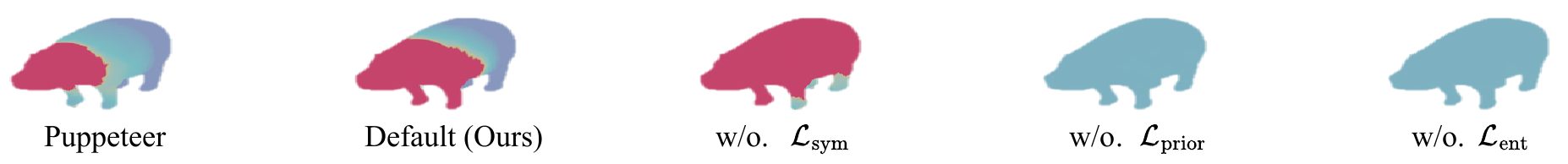}
\caption{
        \textbf{Qualitative skinning ablations.} 
Removing the symmetry loss $\mathcal{L}_{\mathrm{sym}}$ causes weight leakage and violates anatomical structure. Omitting the geometric prior $\mathcal{L}_{\mathrm{prior}}$ or entropy loss $\mathcal{L}_{\mathrm{ent}}$ produces diffuse, low-confidence weights. Our full model yields sharp, localized, and structurally sound skinning, matching or surpassing the pretrained Puppeteer.
    }
    \label{fig:skinning_ablation_1x5}
\end{figure*}

\noindent\textbf{Ablations.}
To validate the necessity of each component in our skinning loss, we perform an ablation study (see \cref{tab:skinning_ablation_vertical}) by removing individual terms while keeping other settings fixed. All model variants converge within $\sim$10 epochs, but removing the distillation terms or regularizers significantly degrades temporal stability and deformation fidelity. Qualitative results (see \cref{fig:skinning-ablation-entropy,fig:skinning_ablation_1x5}) show these components are critical for stable, sharp skinning assignments. Removing the symmetry loss $\mathcal{L}_{\mathrm{sym}}$ (\cref{eq:skl}) causes severe weight leakage, while omitting the geometric prior $\mathcal{L}_{\mathrm{prior}}$ (\cref{eq:prior}) or entropy loss $\mathcal{L}_{\mathrm{ent}}$ (\cref{eq:ent}) produces diffuse, low-confidence weights. We highlight that the entropy regularization (see \cref{fig:skinning-ablation-entropy}) further encourages sparse, confident assignments; without $\mathcal{L}_{\mathrm{ent}}$, weights become ambiguous, reducing stability. We also validate that our automatic anchor frame strategy improves skinning weight generation (see \cref{tab:anchor_definition}, Bottom), outperforming the naive approach of setting $c=1$.

\section{Discussion}
Our work introduces a new perspective on automatic data-driven model rigging by leveraging self-supervised unlabeled dynamic mesh sequences. We propose a general fine-tuning framework that addresses severe temporal inconsistencies in SOTA static rigging models applied to animated sequences. Extensive experiments show that our method achieves superior temporal stability on both skeleton and skinning generation tasks, while maintaining or even improving the model’s original static generation quality. We hope this work serves as a first step toward deformation-driven rigging and inspires further research in this direction.

\noindent \textbf{Future work.} Our framework opens several research directions, such as fine-tuning on longer mesh sequences to capture complex dynamics and improve long-term temporal consistency.


%
%
\bibliographystyle{splncs04}
\bibliography{main}

\end{document}


\title{SPRig: Self-Supervised Pose-Invariant Rigging from Dynamic Mesh Sequences} 

\titlerunning{Abbreviated paper title}

\author{First Author\inst{1}\orcidlink{0000-1111-2222-3333} \and
Second Author\inst{2,3}\orcidlink{1111-2222-3333-4444} \and
Third Author\inst{3}\orcidlink{2222--3333-4444-5555}}

\authorrunning{F.~Author et al.}

\institute{Princeton University, Princeton NJ 08544, USA \and
Springer Heidelberg, Tiergartenstr.~17, 69121 Heidelberg, Germany
\email{lncs@springer.com}\\
\url{http://www.springer.com/gp/computer-science/lncs} \and
ABC Institute, Rupert-Karls-University Heidelberg, Heidelberg, Germany\\
\email{\{abc,lncs\}@uni-heidelberg.de}}

\maketitlesupplementary

\appendix
\setcounter{table}{0}
\renewcommand{\thetable}{\arabic{table}}
\setcounter{figure}{0}
\renewcommand{\thefigure}{\arabic{figure}}

\section{Evaluation Details: LBS RMSE Computation}
\label{sec:supp_lbs_rmse}

Evaluating the LBS RMSE requires per-bone transformations for every frame. Since our framework outputs a pose-invariant static rig (canonical skeleton and skinning weights) without explicit temporal kinematics, we derive the per-frame joint transformations via a differentiable pose fitting procedure. 

Specifically, given the canonical mesh vertices $V_{\mathrm{rest}}$, the predicted skeleton, and the pose-invariant skinning weights $W$, we optimize the local joint rotations (parameterized as axis-angles) and translations for each frame $t$. The objective is to minimize the pointwise $L_2$ distance between the LBS-driven vertices $\hat{V}_t$ and the ground-truth target mesh vertices $V_t$. 

For computational efficiency, we randomly sample 5,000 vertices per frame to compute the loss. The optimization is solved using the Adam optimizer with a learning rate of 0.01 for 60 iterations per frame. To maintain temporal coherence and accelerate convergence, we employ a warm-start strategy: the kinematic parameters optimized for frame $t-1$ are used as the initialization for frame $t$. The final LBS RMSE is then computed across all vertices using these optimally fitted transformations, ensuring the metric strictly reflects the deformation fidelity of the predicted rigs.

\section{Robustness to Extreme Poses and Self-Occlusions}
\label{sec:supp_extreme_poses}

To demonstrate that SPRig learns a generalizable, pose-invariant structural prior rather than trivially memorizing a sequence-specific anchor, we provide qualitative results on challenging out-of-domain static meshes. 

\begin{figure}[htb!]
    \centering
    \includegraphics[width=0.8\linewidth]{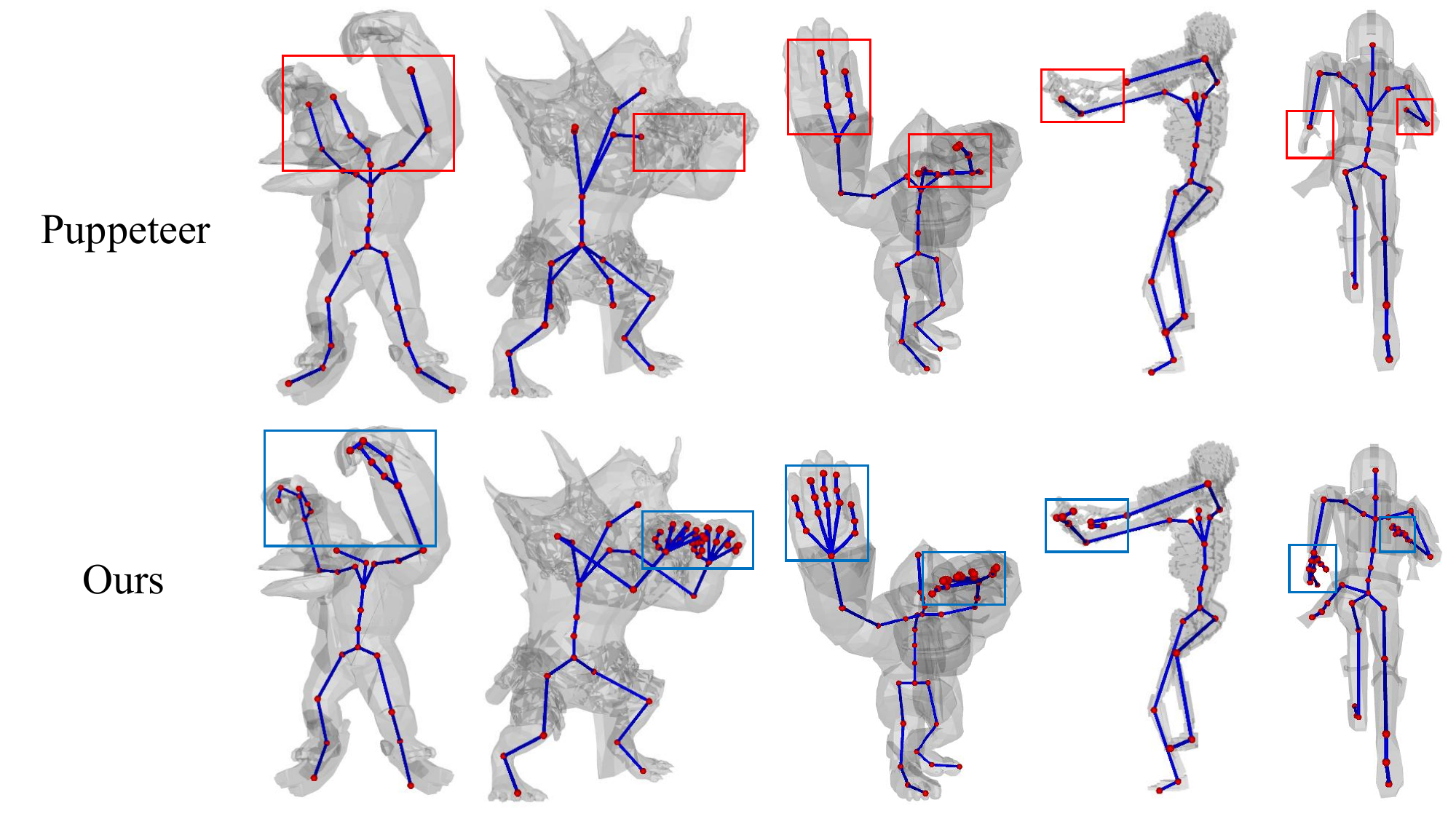}
    \caption{\textbf{Robustness to extreme poses and self-occlusions.} Qualitative comparison between Puppeteer (top) and our SPRig (bottom) on highly articulated character meshes. In challenging poses or under severe self-occlusion, the baseline model frequently drops fine-grained extremities (e.g., missing hand and finger rigs, highlighted in red boxes) because it relies heavily on superficial geometric cues. In contrast, by learning the kinematic invariance of structures during dynamic sequence fine-tuning, SPRig internalizes a robust pose-invariant prior. This enables it to preserve complete, highly detailed skeletal topologies (e.g., full five-finger structures, highlighted in blue boxes) even from single, heavily deformed inputs, demonstrating superior out-of-domain generalization.}
    \label{fig:supp_extreme_poses}
\end{figure}

As shown in Fig.~\ref{fig:supp_extreme_poses}, the baseline (Puppeteer) frequently drops fine-grained distal structures (e.g., individual fingers) under severe self-occlusions. Static models rely heavily on superficial spatial features, which easily collapse when extremities are curled or heavily deformed. 

In contrast, SPRig robustly infers topologically complete skeletons from single, heavily deformed meshes. By enforcing consistency across dynamic frames, where complex structures transition between extended and occluded states, SPRig explicitly learns the underlying \textit{kinematic invariance}. It internalizes that intrinsic skeletal topology persists regardless of severe spatial deformations. This confirms that our framework endows the network with a robust structural prior, enabling superior out-of-domain generalization that extends far beyond simple temporal post-processing.

\section{Failure Cases and Limitations: Symmetric Ambiguity}
\label{sec:supp_failure}

While SPRig significantly improves topological stability and out-of-domain generalization, it can occasionally fail under extreme spatial entanglement—a limitation primarily inherited from the geometric feature extraction mechanism of the underlying static foundation model. 

\begin{figure}[htb!]
    \centering
    \includegraphics[width=0.5\linewidth, trim={0 1cm 0 1cm}, clip]{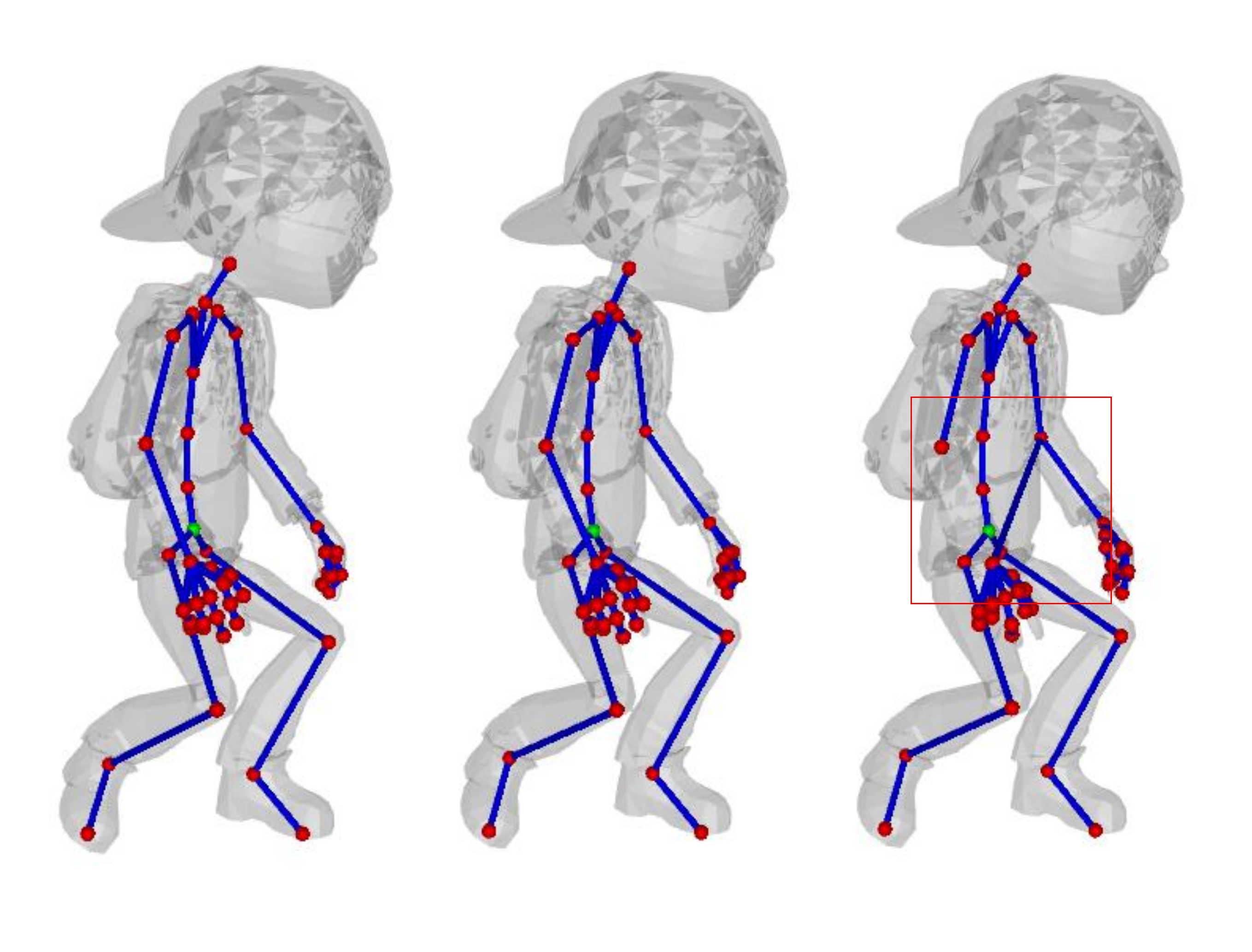} 
    \caption{\textbf{Failure case: Symmetric ambiguity.} During complex poses where distinct parts are in close spatial proximity, the Euclidean-based feature encoder entangles their geometric representations. This ambiguity confuses the autoregressive decoder, causing it to predict an incorrect parent token ($t_{j,p}$). As a result, a spurious connection (a "long bone" artifact highlighted in the red box) bridges kinematically distant but spatially close regions.}
    \label{fig:supp_failure}
\end{figure}

As illustrated in Fig.~\ref{fig:supp_failure}, when symmetric or distinct body parts (e.g., left and right hands, or hands resting on legs) are in extremely close physical proximity during complex interactions, the model suffers from \textit{symmetric ambiguity}. Because the point-cloud encoder of the baseline model aggregates features based on spatial (Euclidean) proximity rather than intrinsic surface geodesics, the geometric signatures of physically touching parts become highly entangled. Consequently, when the autoregressive decoder interprets these entangled features, it is prone to predicting an incorrect \textbf{parent token} ($t_{j,p}$) for the affected joint. Instead of connecting to its true anatomical parent, the joint is erroneously parented to a spatially adjacent but kinematically distant cluster, generating a spurious "long bone" artifact bridging the gap.

Crucially, if such severe spatial entanglement persists even in our automatically selected canonical anchor frame, SPRig's strict consistency mechanism will faithfully—yet incorrectly—propagate this erroneous parent token across the entire sequence. Addressing this limitation is an important direction for future work. Potential solutions include replacing Euclidean-based feature aggregation with geodesic distance metrics (which respect surface topology) or incorporating explicit semantic body-part priors to better disentangle spatially adjacent regions during token decoding.

\section{Video-based Motion Fidelity on the \textit{fish} Sequence}

To quantitatively assess whether the final rig better captures motion, we run the same video-guided animation optimization with two different final rigs: (i) the original Puppeteer rig~\cite{song2025puppeteer} and (ii) our final rig. For both methods, we compare the reconstructed front-view video against the ground-truth driving video. The two videos are first aligned over their common temporal duration, resampled to 10 FPS, and resized to a width of 256 while preserving aspect ratio. The video is attached in the video folder.

We report frame-wise appearance metrics, including L1, PSNR, and SSIM~\cite{wang2004image}, together with two motion-sensitive measures computed from consecutive frames. Specifically, \textit{Temporal-L1} measures the discrepancy between frame-to-frame RGB changes, while \textit{Flow-EPE}$_{\mathrm{norm}}$ measures the normalized endpoint error between dense Farneb\"ack optical flow fields~\cite{farneback2003two}.

As shown in Table~\ref{tab:video_motion_eval_fish}, our method achieves consistent improvements over Puppeteer on the \textit{fish} sequence. In particular, our rig produces lower L1 error (0.147630 vs.\ 0.147960), higher PSNR (10.704060 vs.\ 10.693136), and higher SSIM (0.686730 vs.\ 0.686409). More importantly, it also yields lower temporal discrepancy and lower flow error, with Temporal-L1 reduced from 0.023395 to 0.023275 and normalized Flow-EPE reduced from 0.002342 to 0.002322. These results suggest that our final rig provides a slightly better basis for reconstructing the target motion on this example.

\begin{table}[t]
\centering
\small
\setlength{\tabcolsep}{4pt}
\begin{tabular}{lccccc}
\toprule
Method & L1$\downarrow$ & PSNR$\uparrow$ & SSIM$\uparrow$ & Temporal-L1$\downarrow$ & Flow-EPE$_{\mathrm{norm}}\downarrow$ \\
\midrule
Puppeteer & 0.147960 & 10.693136 & 0.686409 & 0.023395 & 0.002342 \\
Ours       & \textbf{0.147630} & \textbf{10.704060} & \textbf{0.686730} & \textbf{0.023275} & \textbf{0.002322} \\
\bottomrule
\end{tabular}
\caption{Video-based motion fidelity evaluation on the reconstructed front-view video for the \textit{fish} sequence. Lower is better except for PSNR and SSIM.}
\label{tab:video_motion_eval_fish}
\end{table}
\bibliographystyle{splncs04}
\bibliography{main}